%% file: samplepaper.tex
\documentclass[runningheads]{llncs}
\usepackage{amsmath,amssymb}
\usepackage{graphicx}

\usepackage{pgfplots}
\pgfplotsset{compat=1.15}
\usepgfplotslibrary{groupplots}
\usepackage{tikz}
\usepackage{tikzscale}

\newlength\fwidth

\newcommand{\R}{\mathbb{R}}

\DeclareMathOperator*{\argmin}{arg\min}

\DeclareMathOperator*{\Sp}{span}

\begin{document}
\title{Structured Deep Kernel Networks for Data-Driven Closure Terms of Turbulent Flows %
}
\titlerunning{SDKN for Data-Driven Closure Terms of Turbulent Flows}
\author{Tizian Wenzel \inst{1, 5} \and
Marius Kurz\inst{2} \and
Andrea Beck\inst{3} \and
Gabriele Santin \inst{4}\orcidID{0000--0001-6959-1070} \and Bernard Haasdonk \inst{1}}
\authorrunning{T. Wenzel et al.}

\institute{Institute for Applied Analysis and Numerical Simulation,
University of Stuttgart, Germany \and
Institute of Aerodynamics und Gas Dynamics, University of Stuttgart, Germany \and
Laboratory of Fluid Dynamics and Technical Flows, Otto-von-Guericke-Universit\"at Magdeburg \and
Digital Society Center, Bruno
Kessler Foundation, Trento, Italy \and
Corresponding author
}
\maketitle              %
\begin{abstract}
Standard kernel methods for machine learning usually struggle when
dealing with large datasets. We review a recently introduced Structured Deep Kernel
Network (SDKN) approach that is capable of dealing with high-dimensional and huge datasets - and enjoys typical standard machine learning approximation properties. \\
We extend the SDKN to combine it with standard machine learning modules and compare it with Neural Networks on the scientific challenge of data-driven prediction of closure terms of turbulent flows. We show experimentally that the SDKNs are capable of dealing with large datasets and achieve near-perfect accuracy on the given application. %
\keywords{Machine Learning  \and Structured Deep Kernel Networks \and Closure Terms \and Turbulent Flows}
\end{abstract}
\vspace{-1cm}

\input{chapters/introduction}

\input{chapters/machine_learning}

\input{chapters/turbulence_modeling}

\input{chapters/numerical_appl}

\input{chapters/conclusion}

\textbf{Acknowledgements}: 
The authors acknowledge the funding of the project
by the Deutsche Forschungsgemeinschaft (DFG, German Research Foundation)
under Germany’s Excellence Strategy - EXC 2075 - 390740016 and funding by
the BMBF project ML-MORE.
\vspace{-.3cm}

 \bibliographystyle{splncs04}
 \bibliography{references}

\end{document}

%% file: chapters/introduction.tex
\section{Introduction}

In modern science and engineering there is an increasing demand for efficient and reliable data-based techniques. On the one hand, an unprecedented amount of data is nowadays available from various sources, notably complex computer simulations. On the other hand, traditional model-based methods are often required to integrate the additional information acquired through measurements.

A particularly interesting situation is the case of surrogate modeling \cite{SaHa2019a}, where the underlying ground-truth is some engineering model represented by a computational simulation, which is accessible but expensive to execute. In this setting the accurate simulation provides a discrete set of input-output pairs, and the resulting data-based model, or surrogate, can be used to replace to some extent the full simulation in order to simplify its understanding and analysis. %

In this paper we use a recently proposed Structured Deep Kernel Network, which is build on a representer theorem for deep kernel learning \cite{Bohn2017}, and extend and apply it to multivariate regression using time-series data. The technique is applied to the prediction of closure terms for turbulent flows, where a comparison with standard neural network techniques is drawn. The results show the full flexibility of the proposed setup while archiving near-perfect accuracy.

The paper is organized as follows. To begin with, we recall in Section \ref{subsec:machine_learning} and Section \ref{subsec:anns} some background information about machine learning and Artificial Neural Networks (ANNs). In Section \ref{sec:sdkn} the novel Structured Deep Kernel Network (SDKN) is reviewed and combined with standard machine learning modules. Section \ref{sec:turbulence_modeling} provides background information on our application setting and the need for machine learning techniques. The subsequent Section \ref{sec:num_applic} explains the numerical experiments and their results as well as the practicability of the SDKN. Section \ref{sec:conclusion_outlook} summarizes the results and provides an outlook.

\vspace{-.3cm}

\subsection{Regression in Machine Learning} \label{subsec:machine_learning}

Machine learning for regression tasks is usually posed as an optimization problem. For given data $\mathcal{D} := (x_i, y_i)_{i=1}^n$ with $x_i \subset \R^{d_\text{in}}, y_i \subset \R^{d_\text{out}}$, the goal is to find a function $f$ which approximates $f(x_i) = y_i$. Mathematically speaking, this refers to minimizing a loss functional $\mathcal{L}_D$. In case of a mean-squared error (MSE) loss, this means searching for
\vspace{-.2cm}
\begin{align}
\label{eq:loss}
f^* := \argmin_{f \in \mathcal{H}} \mathcal{L}_D(f), \quad \mathcal{L}_D(f) =  \frac{1}{n} \sum_{i=1}^n \Vert y_i - f(x_i) \Vert_2^2 + \lambda \cdot \mathcal{R}(f)
\end{align}
over a suitable space of functions $\mathcal{H}$, whereby $\mathcal{R}(f)$ is a function-dependent regularization term with regularization parameter $\lambda \in \R$. The space of functions $\mathcal{H}$ is usually parametrized, i.e.\ we have $f(x) = f(x, \theta)$ for some parameters $\theta$. %

We consider here two very popular techniques that define $f$, namely Artificial Neural Networks (ANNs) and kernel methods. ANNs enjoy favorable properties for high dimensional data approximation due to representation learning. The other approach is given by kernel methods, which have a sound theoretical background, however struggle when dealing with large and high-dimensional datasets.

\vspace{-.2cm}

\subsection{Neural Networks} \label{subsec:anns}

ANNs (see e.g.\ \cite{Goodfellow-et-al-2016}) are a prevalent class of ansatz functions in machine learning.
The common feedforward ANN architecture consists of $L$ consecutive layers, whereby layer $l$ transforms its input $x \in \R^{d_{l-1}}$ according to
\begin{equation}
f_l(x) = \sigma \left( W_l x + b_l\right),
\end{equation}
with a weight matrix $W_l \in \R^{d_l \times d_{l-1}}$ and a bias vector $b \in \R^{d_l}$. The weight matrices $\{ W_l \}_{l=1,..,L}$ and bias vectors $\{ b_l \}_{l=1,..,L}$ for all the layers are parameters of the ANN, which are obtained by solving the optimization task in Eq.~\eqref{eq:loss}.
A common choice for the non-linear activation function $\sigma(\cdot)$ is the rectified linear unit~(ReLU, $\sigma(\cdot) = \max(\cdot,0)$), which is applied element-wise.
The output of a layer is then passed as input to the succeeding layer.
The feedforward ANN can thus be written as a concatenation of the individual layers $f(x)=f_L \circ .. \circ f_1(x)$.

In contrast to standard kernel methods which will be recalled in the next section, the basis of the ANN is not determined a priori, but depends on its weights and biases.
ANNs can thus learn a suitable basis automatically from the data they are trained on.
Moreover, the concatenation of non-linear transformations allows the ANN to recombine the non-linear basis functions of the preceeding layers to increasingly complex basis functions in the succeeding layers, which renders ANNs highly suitable for high-dimensional data.

Numerous variants of the feedforward ANN have been proposed for particular kinds of data and applications.
For sequential data, recurrent neural networks (RNNs) have established themselves as state of the art.
A common representative of RNNs is the GRU architecture proposed in~\cite{Cho2014}.

\vspace{-.3cm}

%% file: chapters/machine_learning.tex
\section{Structured Deep Kernel networks} \label{sec:sdkn}

\vspace{-.1cm}

Here, we present a brief summary of kernel methods and the architecture presented in \cite{wenzel_sdkn}. For a thorough discussion, we refer the reader to that reference, as our goal here is to present an extension of the orignal SDKN to sequential data.

Standard kernel methods rely on the use of possibly strictly positive definite kernels like the Gaussian kernel
\begin{align}
\label{eq:gaussian}
k(x,z) = \exp(-\Vert x-z \Vert_2^2).
\end{align}
In this framework, a standard Representer Theorem \cite{Wahba1970} simplifies the loss minimization of Eq.\ \eqref{eq:loss} and states that a loss-minimizing function can be found in a subspace $V_n := \Sp \{k(\cdot, x_i), i = 1, .., n \}$ spanned by the data, i.e.\
\begin{align}
\label{eq:shallow_repr_th}
f^*(\cdot) = \textstyle \sum_{j=1}^n \alpha_j k(\cdot, x_j).
\end{align}
The corresponding coefficients $\alpha_j \in \R^{d_\text{out}}$, $j=1, .., n$ can then be found by solving a finite dimensional and convex optimization problem. This is the basis to provide both efficient solution algorithms and approximation-theoretical results. However the choice of a fixed kernel $k(x,y)$ restricts these standard kernel methods, as the feature representation of the data is implicitly determined by the choice of $k$. Furthermore, assembling and solving the linear system to determine the coefficients $\alpha_j$ poses further problems in the big data regime. \\
In order to overcome these shortcoming, a deep kernel representer theorem \cite{Bohn2017} has been leveraged in \cite{wenzel_sdkn} to introduce Strucutured Deep Kernel Networks, which alleviate these obstacles by putting kernel methods into a structured multilayer setup: The deep kernel representer theorem considers a concatenation of $L$ functions as $f = f_L \circ .. \circ f_1$ and states that the problem can be restricted to 
\begin{align}
\label{eq:deep_repr_th}
f_l^*(\cdot) = \textstyle \sum_{j=1}^n \alpha_{lj} k_l(\cdot, f^*_{l-1} \circ .. \circ f^*_1(x_j)), \qquad l = 2, .., L
\end{align}
with $\alpha_{lj} \in \R^{d_l}$. This is a generalization of the standard representation of Eq.\ \eqref{eq:shallow_repr_th} to the multilayer setting. The paper \cite{wenzel_sdkn} proceeds by choosing non-standard kernels: 
\vspace{-.3cm}
\begin{enumerate}
\item For $l$ odd, vector-valued linear kernels are picked: $k_\mathrm{lin}(x,z) = \langle x, z \rangle_{\R^{d_{l-1}}}\cdot I_{d_{l}}$, whereby $I_{d_{l}} \in \R^{d_l \times d_l}$ is the identity matrix.
\item For $l$ even, single-dimensional kernels are used, that use single components $x^{(i)}, z^{(i)}$ of the vectors $x, z \in \R^{d_{l-1}}$: $k_\mathrm{s}(x, z) = \mathrm{diag}(k(x^{(1)}, z^{(1)}), .., k(x^{(d)}, z^{(d)}))$ for some standard kernel $k: \R \times \R \rightarrow \R$, e.g.\ the Gaussian from Eq.\ \eqref{eq:gaussian}.
\end{enumerate}
This choice of kernels introduces a structure which is depicted in Figure \ref{fig:visualization_SDKN}, which motivates the naming \textit{Structured} Deep Kernel Networks and allows for comparison with neural networks: The linear kernels of the odd layers give rise to fully connected layers (without biases), while even layers with their single-dimensional kernels can be viewed as \textit{optimizable activation function layers}. In order to obtain a sparse model and alleviate possible overfitting issues, we only use an expansion size of e.g.\ $M = 5 \ll n$ within the sum of Eq.\ \eqref{eq:deep_repr_th}, which amounts to fix the remaining coefficients to zero. Even for this sparse representation and special choice of kernels, it is proven in \cite{wenzel_sdkn} that the SDKNs satisfy universal approximation properties, legitimizing their use for machine learning tasks. %

The proposed setup is sufficiently flexible such that it can be combined with standard neural network modules: For the application described in Section \ref{sec:turbulence_modeling} we incorporate GRU-modules by using them in between two activation function layers. By doing so, we can make use of the time-dependence within the input data.

\vspace{-.6cm}

\input{Figures/tikzfigures/figures_SDKN.tex}

\vspace{-1cm}

%% file: Figures/tikzfigures/figures_SDKN.tex
\begin{figure}%
\def\layersep{2.2cm}
\centering %
\begin{tikzpicture}[scale=0.75, shorten >=1pt,->,draw=black!50, node distance=\layersep]
    \tikzstyle{every pin edge}=[<-,shorten <=1pt]
    \tikzstyle{neuron}=[circle,fill=black!25,minimum size=8pt,inner sep=0pt]
    \tikzstyle{input neuron}=[neuron, fill=green!50];
    \tikzstyle{output neuron}=[neuron, fill=red!50];
    \tikzstyle{hidden neuron}=[neuron, fill=blue!50];
    \tikzstyle{annot} = [text width=5em, text centered]

	\node[annot] () at (-1.5, -2.5) {Input};
	\node[annot] () at (5.75*\layersep, -2.5) {Output};

    \foreach \name / \y in {1,2,3}
        \node[input neuron,pin={[pin edge={-}]left:}] (I-\name) at (0,-.5-\y) {};

    \foreach \name / \y in {1,...,4}
        \node[hidden neuron] (H1-\name) at (\layersep,-\y) {};
        
    \foreach \name / \y in {1,...,4}
        \node[hidden neuron] (H2-\name) at (2*\layersep,-\y) {};
        
    \foreach \name / \y in {1,...,4}
        \node[hidden neuron] (H3-\name) at (3*\layersep,-\y) {};
        
    \foreach \name / \y in {1,...,4}
        \node[hidden neuron] (H4-\name) at (4*\layersep,-\y) {};

    \foreach \name / \y in {1,2}
        \node[output neuron,pin={[pin edge={-}]right:}] (O-\name) at (5*\layersep,-1-\y) {};

    \foreach \source in {1,...,3}
	    \foreach \dest in {1,...,4}
	        \path [lightgray] (I-\source) edge (H1-\dest);    
    
    \foreach \source in {1,...,4}
        \foreach \dest in {1,...,4}
            \path [lightgray] (H2-\source) edge (H3-\dest);
            
	\foreach \source in {1,...,4}
		\foreach \dest in {1,2}
	        \path [lightgray] (H4-\source) edge (O-\dest);   

	\foreach \node in {1,...,4}
		\path [black] (H1-\node) edge (H2-\node);
		
	\foreach \node in {1,...,4}
		\path [black] (H3-\node) edge (H4-\node);

	\node[annot] (hl1) at (.9, -.2) {$f_1$ using kernel $k_\text{lin}$};
    \node[annot] (hl2) at (3.2, -.2) {$f_2$ using kernel $k_s$};
    \node[annot] (hl3) at (5.4, -.2) {$f_3$ using kernel $k_\text{lin}$};
	\node[annot] (hl4) at (7.6, -.2) {$f_4$ using kernel $k_s$};
	\node[annot] (hl5) at (9.8, -.2) {$f_5$ using kernel $k_\text{lin}$};

	\draw[decorate, thick, -, decoration={brace,mirror}, yshift=3ex]  (.1*\layersep, -5) -- node[below=2ex, align=center, midway] {fully \\ connect \\ layer}  (.9*\layersep, -5);
	\draw[decorate, thick, -, decoration={brace,mirror}, yshift=3ex]  (1.1*\layersep, -5) -- node[below=2ex, align=center, midway] {activation \\ function \\ layer}  (1.9*\layersep, -5);
	\draw[decorate, thick, -, decoration={brace,mirror}, yshift=3ex]  (2.1*\layersep, -5) -- node[below=2ex, align=center, midway] {fully \\ connect \\ layer}  (2.9*\layersep, -5);	
	\draw[decorate, thick, -, decoration={brace,mirror}, yshift=3ex]  (3.1*\layersep, -5) -- node[below=2ex, align=center, midway] {activation \\ function \\ layer}  (3.9*\layersep, -5);	
	\draw[decorate, thick, -, decoration={brace,mirror}, yshift=3ex]  (4.1*\layersep, -5) -- node[below=2ex, align=center, midway] {fully \\ connect \\ layer}  (4.9*\layersep, -5);	

\end{tikzpicture}
\caption{Visualization of the Structured Deep Kernel Network. Gray arrows refer to layers using the linear kernel, while black arrows refer to layers using the single dimensional kernel layer. The braces below the layers indicate similarities to neural networks.}
\label{fig:visualization_SDKN}
\end{figure}

%% file: chapters/turbulence_modeling.tex
\section{Turbulence Closure Problem}
\label{sec:turbulence_modeling}

\vspace{-.1cm}

The evolution of compressible fluid flows (see e.g.\ \cite{Serrin1959}) is governed by the Navier-Stokes equations, which can be written in short as
\begin{equation}
  U_t + R(F(U)) = 0 ,
  \label{eq:discrete_nse}
\end{equation}
with $U$ as the vector of conserved variables.
$U_t$ denotes the derivation with respect to time and~$R()$ denotes the divergence operator applied to the non-linear fluxes~$F(U)$.
Most engineering flows of interest exhibit turbulent behavior.
Such turbulent flows are inherently chaotic dynamical systems with a multiscale character.
The direct numerical simulation (DNS) of turbulence is thus only feasible for simple geometries and low Reynolds numbers, which correspond to a small range of active flow scales.
The framework of large eddy simulation (LES) addresses these restrictions by resolving only the large energy-containing scales by applying a low-pass filter $\overline{(\cdot)}$ to the underlying Navier-Stokes equations.
However, the filtered flux term~$\overline{R(F(U))}$ is generally unknown, since it depends on the full solution~$U$.
To this end, the coarse-scale solution is typically advanced in time by some appropriate numerical discretization $\tilde{R}(\overline{U})$, which then yields
\begin{equation}
  \overline{U}_t + \tilde{R}(\overline{U}) = \underbrace{\tilde{R}(\overline{U}) - \overline{R(F(U))}}_{\mathrm{perfect\:LES\:closure}}.
  \label{eq:perfect_les}
\end{equation}
Solving these filtered equations becomes feasible in terms of computational cost, but the right-hand side of Eq.~\eqref{eq:perfect_les} exhibits the unknown \textit{closure terms}, which describe the effects of the unresolved small-scale dynamics on the resolved scales.

The task of turbulence modeling can thus be stated as finding some model $M$ which recovers the closure terms solely from the filtered flow field:
\vspace{-.1cm}
\begin{equation}
  \left(\tilde{R}(\overline{U}) - \overline{R(F(U))}\right) \approx M(\overline{U}) .
  \label{eq:turbulence_modeling}
\end{equation}
A myriad of different models have been proposed in literature over the last decades to derive the mapping~$M$ based on mathematical and physical reasoning.
While the accuracy of the closure model is crucial for obtaining a reliable description of the flow field, no universal and generally \textit{best} model has been identified to date.
Therefore, increasing focus is laid upon finding this mapping~$M$ from data by leveraging the recent advances in machine learning.
See~\cite{beck2021perspective} for an extensive review.
In that reference, machine learning is used to directly recover the unknown flux term $\overline{R(F(U))}=f(\overline{U})$ without positing any prior assumptions on the functional form of the underlying mapping~$f(\cdot)$.

\vspace{-.3cm}

%% file: chapters/numerical_appl.tex
\section{Numerical application} \label{sec:num_applic}
\begin{figure}[t]
  \centering
  \input{Figures/dataset_data/fig_filter_field_solution.tikz}
  \caption{Two-dimensional slices of the three-dimensional z-velocity field $v^{(3)}$ for the high-fidelity DNS and for the corresponding filtered velocity fields $\overline{v}^{(3)}$ of the three investigated LES filters. Each slice of the filtered flow field contains $48^2$ solution points.}
  \label{fig:training_data}
\end{figure}
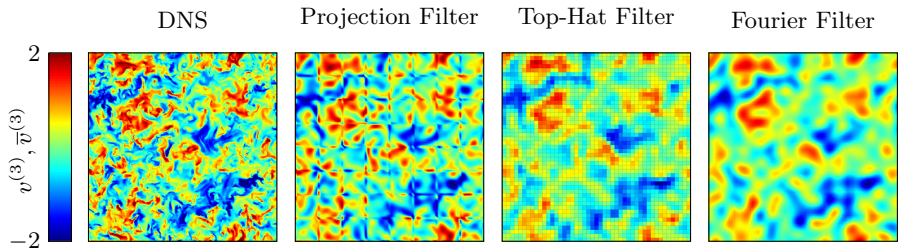

In the present work, the dataset described in \cite{beck2019deep,kurz2020machine} (to which we refer for further details on the following configurations) is used as training set for the machine learning algorithms.
This dataset is based on high-fidelity DNS of decaying homogeneous isotropic turbulence~(DHIT).
The coarse-scale quantities according to Eq.~\eqref{eq:perfect_les} are obtained by applying three different LES filters to the DNS solution: A global Fourier cutoff filter (``Fourier''), a local $L_2$-projection filter onto piecewise polynomials (``projection'') and a local Top-hat filter, as shown in Figure~\ref{fig:training_data}.
The latter two are based on the typical representations of the solution in discontinuous Galerkin (DG) and Finite-Volume (FV) schemes, respectively.

For each filter, the corresponding dataset comprises nearly 30 million samples and a separate DHIT simulation is used as blind testing set.
Since turbulence is a time-dependent phenomenon, the input features for each training sample are chosen as a time series of the filtered three-dimensional velocity vector $\overline{v}^{(i)}$, $i=1,2,3$ at a given point in space.
The target quantity is the three-dimensional closure term $\overline{R(F(U))}{}^{(i)}$, $i=1,2,3$ in the last timestep of the given series.
To investigate the influence of temporal resolution, a variety of different sampling strategies are examined, which are given in Table~\ref{tab:nomenclature_GRU}.
Before training, the input features of the training dataset were normalized to zero mean and unit variance.
\vspace{-.8cm}
\begin{table}
  \centering
  \def\arraystretch{1.1}
  \caption{Different time series for training. $N_{seq}$ denotes the number of time instances per sample and $\Delta t_{seq}$ is the time increment between two successive time instances.}
  \begin{tabular}{lclllll}
    \hline
                     && GRU1      && GRU2      && GRU3 \\
    \hline
    $N_{seq}$        &&       $3$ &&      $10$ &&      $21$ \\
    $\Delta t_{seq}$ && $10^{-3}$ && $10^{-4}$ && $10^{-4}$ \\
    \hline
  \end{tabular}
  \label{tab:nomenclature_GRU}
\end{table}
\vspace{-.3cm}

\begin{figure}[ht]
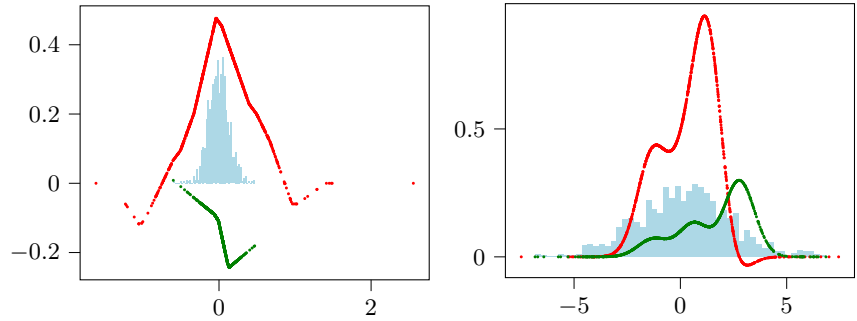

\setlength\fwidth{.4\textwidth}
\centering
\input{Figures/tikzfigures/fig_dim14}
\input{Figures/tikzfigures/fig_dim2}
\caption{Visualization of exemplary single-dimensional kernel function mappings before (red) and after optimization (green). The histogramms indicate the distribution of the training data after optimization. Right: Gaussian kernel as defined in Eq.\ \eqref{eq:gaussian}. Left: Wendland kernel of order $0$, which is defined as $k(x,y) = \max(1-\Vert x - y \Vert_2, 0)$. These mappings can be interpreted as optimizable activation functions, see Section \ref{sec:sdkn}.}
\label{fig:testfig}
\end{figure}

\vspace{-.4cm}
The ANN architecture from~\cite{kurz2020machine} was used as baseline model, which incorporates a GRU layer to leverage the temporal information in the data.
Such a GRU layer was also introduced into the SDKN framework, which clearly demonstrates its modularity.
In order to obtain a fair comparison of both the SDKN and ANN approach, both models exhibit mostly the same setup: The structure of the networks is given via their input dimension and output dimension of $3$ each and the dimensions of the inner layers are chosen as $32, 64, 48, 24$.
This amounts to a total of $31400$ optimizable parameters for the ANN and $32240$ for the SDKN.
The slight difference is related to the additional parameters $\alpha_{lj}$ within the single-dimensional kernel layers, see Eq.\ \eqref{eq:deep_repr_th} for $l$ even, and the unused bias-parameters within the SDKN. Concerning the the results in Table \ref{tab:results_marius_corr}, the Gaussian kernel from Eq.\ \eqref{eq:gaussian} was used for the single-dimensional kernels within the SDKN, but we remark that the use of other kernels yields qualitatively the same results.
We further remark that both models are within the underparametrized regime, as the number of parameters is significantly below the number of training samples.

The Adam optimizer was used for training with an initial learning rate of $10^{-3}$, which was then halved after each 10 epochs for the NN and after each 5 epochs for the SDKN.
The NN was optimized for $50$ epochs and the SDKN for $25$ epochs.
This distinction keeps the overall optimization time comparable, since the SDKN optimization is about a factor of $2$ more time-consuming per epoch.
This stems from the optimization of the additional parameters (e.g.\ the $\alpha_{lj}$ for $l$ even in Eq.\ \eqref{eq:deep_repr_th}) related to its setup.
For training, the MSE loss from Eq.\ \eqref{eq:loss} was used without any regularization, since the use of a relatively small batch size (128) is likely to provide a sufficient regularization effect.
We remark that both architecture and training are hyperparameter-optimized for the ANN setting, as this was the model of choice in \cite{kurz2020machine}.
The experiments were run on an Nvidia GTX1070 using PyTorch implementations of the models. The optimization took about 12 GPU-hours each.

Table \ref{tab:results_marius_corr} summarizes the results for the different datasets and cases: Both the final MSE-loss as well as the cross-correlation is given.
The reported cross-correlations match the ones reported in \cite{kurz2020machine}, thus validating their results: Even in our runs, the ANN using the GRU2 setup performed badly on the DG dataset with only a final cross-correlation of $0.8163$, which is way worse than the other listed cross-correlations. The SDKN reached at least the same test accuracies, even without any further hyperparameter optimization. Especially there is no drop in accuracy for the prediction related to the DG dataset in conjunction with the GRU2 case, as it can be observed in Table \ref{tab:results_marius_corr}: The SDKN reaches a cross-correlation of $0.9989$ which is en par with the other GRU-setups, in contrast to the performance of the ANN setup. This might indicate that the SDKN is less likely to get stuck in a local minima compared to the ANN.
\vspace{-.3cm}

\begin{table}[h]
\centering
\caption{Overview of the results on the test set after optimization of the Neural Network (NN) and the Structured Deep Kernel Network (SDKN): Cross-correlation (left) and Loss (right).} 
\renewcommand{\arraystretch}{1.4}
\tiny
\begin{tabular}{|ll|r|r|l|l|} \hline
\textbf{Cross-Correlation} 	&			& GRU1 		& GRU2 		& GRU3   \\ \hline \hline
Projection             	& ANN		& 0.9989 	& 0.8163	& 0.9989 \\ 
			               	& SDKN		& 0.9989 	& 0.9989 	& 0.9988 \\ \hline
Top-Hat                	& ANN		& 0.9992 	& 0.9992  	& 0.9992 \\ 
			               	& SDKN		& 0.9991	& 0.9992 	& 0.9992 \\ \hline
Fourier \qquad \qquad		& ANN		& 0.9992	& 0.9993   	& 0.9993 \\ 
			               	& SDKN		& 0.9992 	& 0.9993	& 0.9993 \\ \hline
\end{tabular}
\quad 
\begin{tabular}{|ll|r|r|l|l|} \hline
\textbf{MSE-Loss} 		   	&			& GRU1 			& GRU2 			& GRU3   \\ \hline \hline
Projection           	& ANN		& 3.235e-01 	& 4.996e+01		& 3.233e-01 \\ 
			           	& SDKN		& 3.253e-01 	& 3.261e-01 	& 3.368e-01 \\ \hline
Top-Hat             	& ANN		& 3.155e-02 	& 2.917e-02  	& 2.888e-02 \\ 
			           	& SDKN		& 3.222e-02		& 2.989e-02 	& 2.893e-02 \\ \hline
Fourier \qquad \qquad	& ANN		& 1.179e-02		& 9.737e-03    	& 9.452e-03 \\ 
			           	& SDKN		& 1.177e-02 	& 1.007e-02		& 9.587e-03 \\ \hline
\end{tabular}
\label{tab:results_marius_corr}
\end{table}

%% file: Figures/dataset_data/fig_filter_field_solution.tikz
\def\XIndex{0}  
\def\YIndex{1}  
\def\DataA{3}   
\def\DataB{6}   
\def\DataC{9}   
\def\figwidth{0.335\textwidth} 

\def\DataAMin{-2}    
\def\DataAMax{+2}    
\def\DataBMin{-20}   
\def\DataBMax{+20}   
\def\DataCMin{-60}   
\def\DataCMax{+60}   

\pgfplotsset{/pgfplots/colormap={jet}{rgb255(0cm)=(0,0,128)rgb255(1cm)=(0,0,255)rgb255(3cm)=(0,255,255)rgb255(5cm)=(255,255,0)rgb255(7cm)=(255,0,0)rgb255(8cm)=(128,0,0)}}

\begin{tikzpicture}[font=\small,baseline]

  \begin{groupplot}[
        group style={
          group size=4 by 1,
          horizontal sep={0.02\textwidth},
        },
        width=\linewidth
    ]

    \nextgroupplot[
      title={DNS},
      axis on top,
      enlargelimits=false,
      width={\figwidth},height={\figwidth},
      ticks=none,
      point meta min=\DataAMin,
      point meta max=\DataAMax,
      colorbar,
      colorbar left,
      colorbar style={
        samples=11,
        width=0.3cm,
        title={$v^{(3)},\overline{v}^{(3)}$},
        ylabel near ticks,
        title style={at={(-0.20,0.4)},anchor=south,rotate=90},
        ytick={-2,2},
        at={(-0.15,0.)},
        anchor=south,
      },
      view={0}{90}]
      \addplot graphics[xmin=0,ymin=0,xmax=6.282,ymax=6.282] {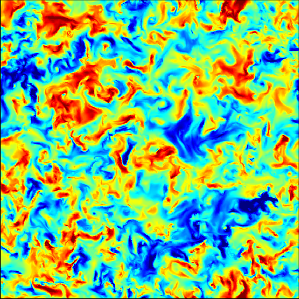};

    \nextgroupplot[
      title={Projection Filter},
      axis on top,
      enlargelimits=false,
      width={\figwidth},height={\figwidth},
      ticks=none,
      point meta min=\DataAMin,
      point meta max=\DataAMax,
      view={0}{90}]
      \addplot graphics[xmin=0,ymin=0,xmax=6.282,ymax=6.282] {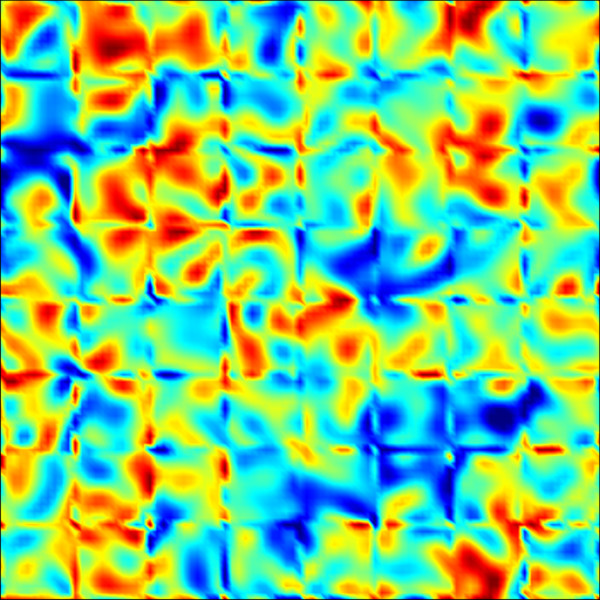};

    \nextgroupplot[
      title={Top-Hat Filter},
      axis on top,
      enlargelimits=false,
      width={\figwidth},height={\figwidth},
      ticks=none,
      point meta min=\DataAMin,
      point meta max=\DataAMax,
      view={0}{90}]
      \addplot graphics[xmin=0,ymin=0,xmax=6.282,ymax=6.282] {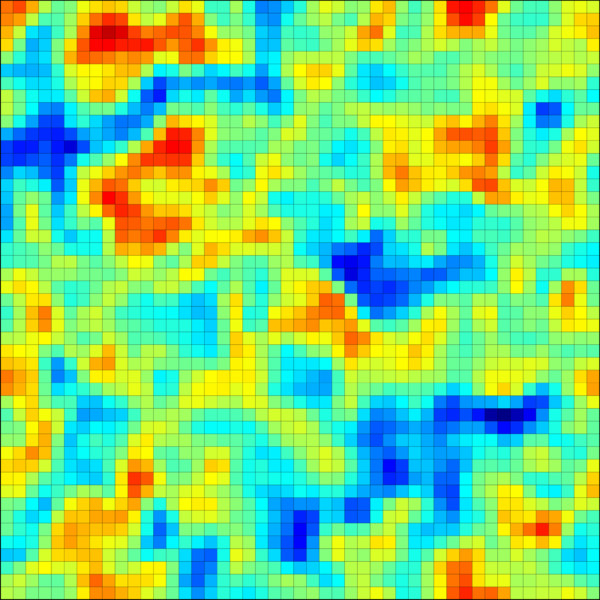};

    \nextgroupplot[
      title={Fourier Filter},
      axis on top,
      enlargelimits=false,
      width={\figwidth},height={\figwidth},
      ticks=none,
      point meta min=\DataAMin,
      point meta max=\DataAMax,
      view={0}{90}]
      \addplot graphics[xmin=0,ymin=0,xmax=6.282,ymax=6.282] {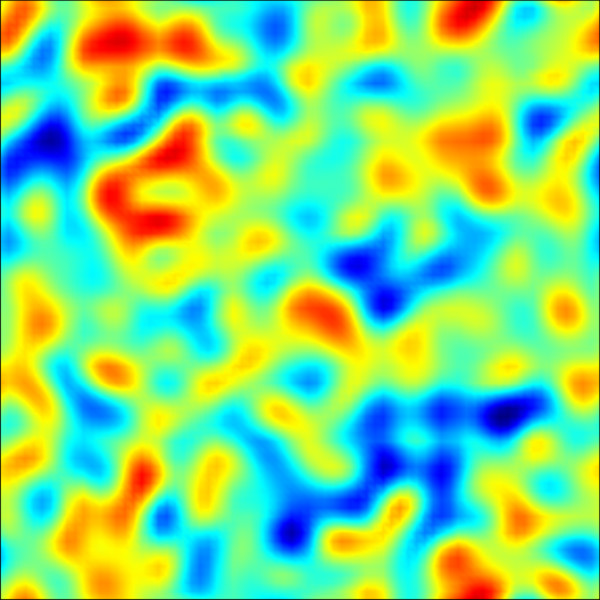};

  \end{groupplot}
\end{tikzpicture}

%% file: chapters/conclusion.tex
\section{Conclusion and outlook} \label{sec:conclusion_outlook}

\vspace{-.1cm}

In this paper an extension of a recently proposed Structured Deep Kernel Network (SDKN) setup to time-series data was introduced. The proposed SDKN model was compared to standard Neural Network models on the challenging task of data-driven prediction of closure terms for turbulence modeling. With help of machine learning models, significant speed ups in the simulation of turbulent flows can be achieved. It was shown numerically that the SDKN can reach the same near-perfect accuracy as hyper-parameter optimized ANNs for several variants of the task at the same training cost. \\
The mathematical background which the SDKN was built on seems promising for further theoretical analysis. From the application point of view, the goal is to optimize the SDKN only for a few epochs and then use the optimized kernel in conjunction with standard shallow kernel models, for which efficient and reliable methods exists. This is expected to significantly speed up the training phase of the surrogate model.